%% file: paper.tex
\pdfoutput=1

\documentclass[runningheads]{llncs}

\usepackage{import}
\usepackage{graphicx}
\usepackage{comment}
\usepackage{subfig}
\usepackage{amsmath,amssymb} % define this before the line numbering.
\usepackage{color}
\usepackage{booktabs}
\usepackage{multirow}
\usepackage{gensymb}

\begin{document}

\pagestyle{headings}
\mainmatter
\def\ECCVSubNumber{3794}  % Insert your submission number here

%-------------------------------------------------------------------------
% Title
% \title{Monocular Differentiable Rendering for Self-Supervised 3D Object Detection}
% \title{Supplementary Material}
% \titlerunning{ECCV-20 submission ID \ECCVSubNumber}
% \authorrunning{ECCV-20 submission ID \ECCVSubNumber}

% \institute{Paper ID \ECCVSubNumber}

% CAMERA READY SUBMISSION
%\begin{comment}
\title{Monocular Differentiable Rendering for Self-Supervised 3D Object Detection}
% \title{Supplementary Material}
\author{Deniz Beker\inst{1}, Hiroharu Kato\inst{1}, Mihai Adrian Morariu\inst{1}, Takahiro Ando\inst{1}, Toru Matsuoka\inst{1}, Wadim Kehl\inst{2}, Adrien Gaidon\inst{3}}
\authorrunning{Beker et al.}
\institute{Preferred Networks, Inc  \and %
           Toyota Research Institute - Advanced Development \and 
           Toyota Research Institute}
\maketitle
\thispagestyle{empty}

\index{Morariu, Mihai}

\titlerunning{Monocular Differentiable Rendering for Self-Supervised 3D Object Detection}

%-------------------------------------------------------------------------
% Abstract
\input{00_abstract.tex}

%-------------------------------------------------------------------------

\input{01_introduction.tex}
\input{02_related_work.tex}

\input{03_method.tex}
\input{05_experiments}
\input{06_conclusion}
\input{07_appendix.tex}
\input{08_acknowledgements}

%-------------------------------------------------------------------------
%\clearpage
{
    %\small
    % \bibliographystyle{ieee_fullname}
    % \bibliography{egbib}
}

%-------------------------------------------------------------------------
% \clearpage
%\appendix

\small
\bibliographystyle{ieee_fullname}
\bibliography{paper}

\end{document}

%% file: 00_abstract.tex
\begin{abstract}

3D object detection from monocular images is an ill-posed problem due to the projective entanglement of depth and scale. To overcome this ambiguity, we present a novel self-supervised method for textured 3D shape reconstruction and pose estimation of rigid objects with the help of strong shape priors and 2D instance masks. Our method predicts the 3D location and meshes of each object in an image using differentiable rendering and a self-supervised objective derived from a pretrained monocular depth estimation network.
We use the KITTI 3D object detection dataset to evaluate the accuracy of the method. Experiments demonstrate that we can effectively use noisy monocular depth and differentiable rendering as an alternative to expensive 3D ground-truth labels or LiDAR information.
\end{abstract}

%% file: 01_introduction.tex
\section{Introduction}
\label{sec:introduction}
% This is where we define the initial problem
Autonomous driving relies heavily on 3D object perception for safe navigation. Most existing systems leverage active sensors (e.g., LiDAR, radar) for location estimation, yet they are either prohibitively costly for large-scale deployment or too sparse in spatial coverage. For these reasons, research in monocular 3D object detection has seen rising popularity in recent years.  

%%%%%%%% figure %%%%%%%%%%%%%%%%%%%%%%%%
\begin{figure*}[t]
    \begin{center}
        \includegraphics[ scale=0.7]{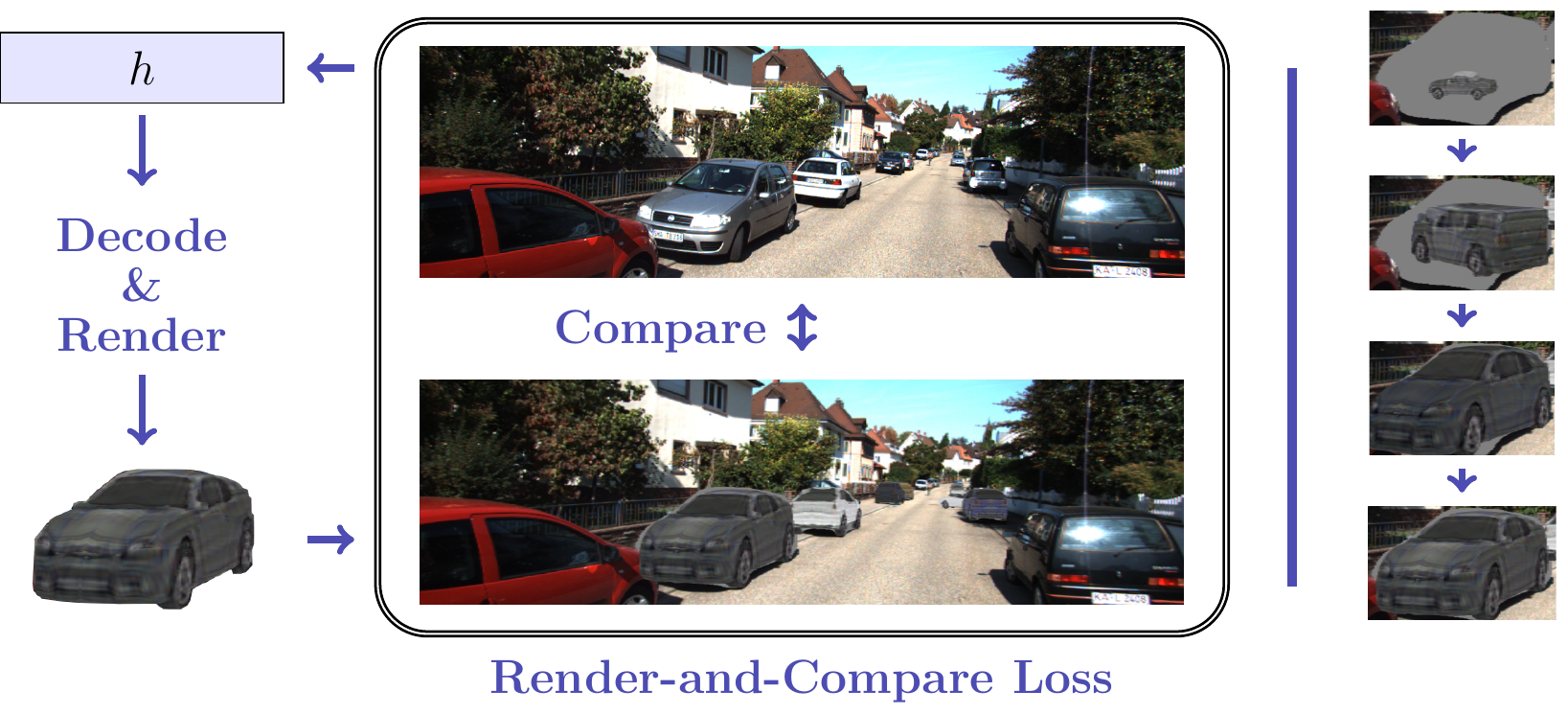}
    \end{center}
    \caption{We propose a self-supervised optimization pipeline for monocular 3D object estimation via analysis-by-synthesis over object pose, metric dimensions, shape and texture. Left: Starting from a 2D instance mask detection, we feed the latent variables $h$ into our decoder network and use differentiable rendering to obtain 2D projections. We use various render-and-compare losses over multiple quantities for comparative analysis and back-propagate the error. Right: The fitting process over multiple iterations. Starting from a random initialization, we can recover the actual object properties quite well.}
    \label{fig:teaser}
\end{figure*}

% This is how the existing solutions tackle and what we want to change differently from existing solutions
Ongoing advancements have led to steadily improving detection accuracy~\cite{chen2016monocular,manhardt2019roi,simonelli2019disentangling}. Despite foregoing active sensing, they still require supervision in the form of 9D cuboid labels which encode 3D location, rotation and metric object dimensions. Most often, such labels are obtained with the help of annotation pipelines and 3D LiDAR point clouds, demanding the usage of costly human labour and expensive sensors. Alternatively, one can leverage self-supervised autolabeling techniques that employ strong shape priors and optimize 3D alignment in stereo point clouds \cite{engelmann2016auto,engelmann2017samp,wang2020autolabeling} or mixed RGB/LiDAR setups \cite{zakharov2020autolabeling}. 

% This is the explanation of how we achieve self-supervision, monocular depth estimation
In this paper, we explore a novel solution towards tackling 3D location and shape estimation from 2D instance mask detections, requiring only monocular input and geometric priors for self-supervision. Recently, self-supervised depth estimation networks~\cite{guizilini20193d,pillai2019superdepth} have become interesting alternatives to LiDAR sensing. They are generally trained using only video sequences and are able to estimate depth with absolute scale when combined with weak supervision (e.g., velocity of the ego-camera~\cite{guizilini20193d}). However, they tend to be rather noisy, overall less accurate than LiDAR, and therefore not well-suited for precise 3D object localization and metric dimension estimation. We therefore incorporate additional strong priors over learned textured object shapes for regularization and run comparative scene analysis via differentiable rendering~\cite{loper2014opendr,kato2018neural,liu2019soft,chen2019diff}. This allows for optimizing 3D variables against image evidence via self-supervision and back-propagation. While initial work focused mostly on toy examples, recent papers have successfully explored applications of analysis-by-synthesis in the wild~\cite{kundu2018cvpr,kulkarni2019csm,zuffi2019safari,zakharov2020autolabeling} and we follow in their path. 

We present an example optimization of our pipeline in Fig.~\ref{fig:teaser}. Starting from a 2D detector that produces 2D boxes with associated instance masks, we optimize over the 3D location, rotation, metric dimensions as well as the object's shape and texture. In our paper, we handle metric dimension as 3D scaling. For the actual alignment computation, we leverage complimentary cues in the form of instance masks, RGB values as well as monocular depth, and regularize with our textured shape space. While our initial estimation is quite rough, our method is able to converge to a good solution while traversing the possible space of shapes, textures, metric sizes and poses. This process is run sequentially until all 2D detections in the scene have been parsed and transformed into 3D estimates.

To summarize our main contributions: Firstly, we remove the necessity for 3D sensing or ground-truth information, making our 3D object estimation pipeline a truly monocular approach. Secondly, we propose the idea of regularizing noisy monocular depth maps at the instance level by strong geometric object priors. Lastly, we evaluate our pipeline on the KITTI 3D dataset~\cite{geiger2012we} and show that we can achieve comparable accuracy to SoTA methods that rely on 3D supervision.

%% file: 02_related_work.tex
\section{Related Work}

Due to the huge body of related 3D detection work, we will focus our survey mostly on monocular methods. Mono3D~\cite{chen2016monocular} is an early work which introduces a region proposal method tailored to autonomous driving. This method computes the 3D bounding box by exploiting different cues such as semantic segmentation, instance segmentation, shape, context features and absolute locations in 2D/3D space. Importantly, they incorporate priors to limit the cuboid search space by using pre-defined object sizes and orientations as well as assuming a known ground plane. Deep3DBox~\cite{mousavian20173d} estimates cuboids from 2D bounding boxes, assuming that the former should be tightly fit to the latter when projected onto the image plane. This strong assumption can underperform in cases where the actual 2D bounding boxes are not tightly enclosing the objects  due to occlusion or truncation. 
The advent of unsupervised/self-supervised monocular depth estimation~\cite{godard2017unsupervised,mahjourian2018depth,pillai2019superdepth} saw their inclusion in recent detection work. In Multi-Fusion~\cite{xu2018multi} 3D region proposals are generated by backprojecting 2D proposals to the 3D space using monocular depth. Similarly, ROI-10D~\cite{manhardt2019roi} employs such depth maps to robustify their bounding box lifting from the 2D to the 3D space.

Methods based on the idea of Pseudo-LiDAR~\cite{wang2019pseudo,ma2019accurate} leverage reprojected monocular depth from off-the-shelf models and run detection networks that have been originally designed for LiDAR input with impressive accuracy improvements. All the mentioned approaches benefit greatly from the employed monocular depth modules but their analysis shows that their overall performance is heavily dominated by the accuracy of depth estimation. For this reason, regularization from additional priors is desirable. 

In terms of analysis-by-synthesis, much novel work has been presented in respect to differentiable rendering. The works \cite{loper2014opendr,kato2018neural,chen2019diff,Li:2018:DMC} propose different ways to produce gradients for the rasterization of triangle meshes. Building on such ideas, both \cite{kundu2018cvpr} as well as \cite{zakharov2020autolabeling} present render-and-compare optimization frameworks with learned shape spaces for automotive scenarios. The former uses 2D instance masks for estimating the shape and up-to-scale pose whereas the latter leverages LiDAR observations for full 3D estimation. 

There exists other slightly-related work that leverages learned shape spaces for automotive object retrieval that we discuss for completeness. In \cite{engelmann2016auto} a detector initializes object instances which are further optimized for pose and shape priors over stereo depth. The authors later extended it~\cite{engelmann2017samp} for temporal priors to also recover pose trajectories. In a similar vein, \cite{wang2020autolabeling} runs full 3D object pose and shape recovery over stereo depth. The authors of \cite{stutz2018completion} explore probabilistic 3D object completion via a shape space and LiDAR as weak supervision. Their work assumes correct localization and focuses solely on the reconstruction quality.

%% file: 03_method.tex
\section{Method}
\begin{figure*}[t]
    \begin{center}
        \includegraphics[ scale=0.35]{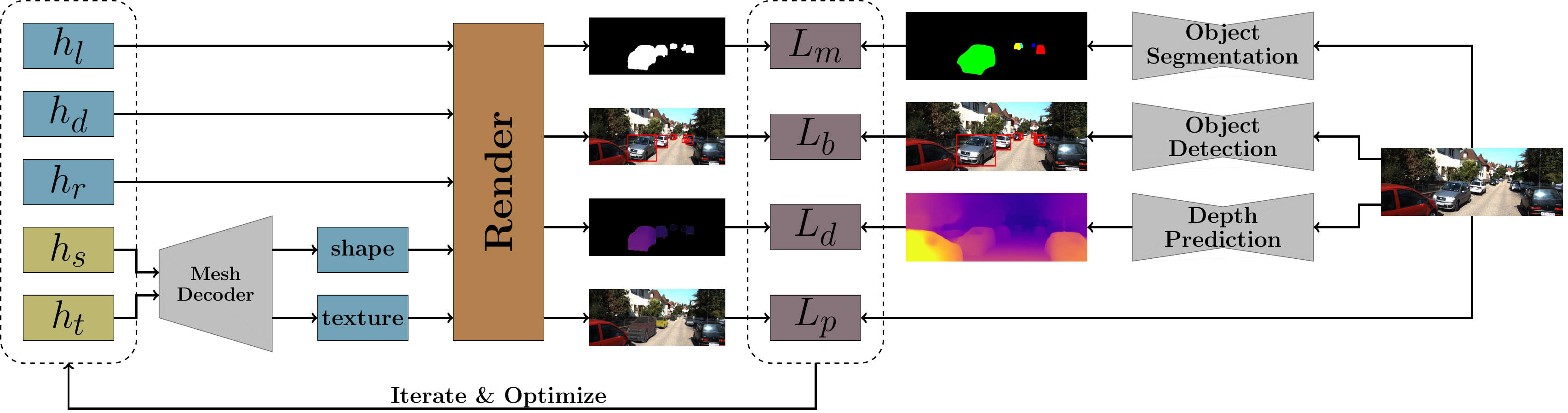}
    \end{center}
    \caption{Overall pipeline of our method where $h_l$, $h_d$, $h_r$ are the variables for location, dimension and rotation, and $h_s$, $h_t$ are the latent variables for shape and texture. We denote with $L_m$, $L_b$, $L_d$, $L_p$ the losses for the silhouette, bounding box, depth map and pixel colors respectively.}
    \label{fig:pipeline}
\end{figure*}

As mentioned, most existing monocular 3D detection methods rely on supervised learning, requiring 3D cuboid information. Instead, we propose to estimate all required 3D object properties via self-supervision and differentiable rendering, avoiding any form of 3D annotation. We depict our pipeline in Fig.~\ref{fig:pipeline} and will provide an overview before going into more detailed descriptions in the next subsections. 

We initially run off-the-shelf object detectors on an image to produce masked 2D detections. For each detection, we place an initial estimate into 3D space by instantiating all object properties. Concretely, we capture metric location $h_{\textrm loc} \in \mathbb{R}^{3}$, metric dimension  $h_{\textrm dim} \in \mathbb{R}^{3}$, rotation along the yaw axis $h_{\textrm rot} \in \mathbb{R}$, shape $h_{\textrm sh} \in \mathbb{R}^{D_{\textrm sh}}$ and texture $h_{\textrm tx} \in \mathbb{R}^{D_{\textrm tx}}$. Following the parameterization introduced in ROI-10D~\cite{manhardt2019roi}, we formulate the location $h_{\textrm loc}$ as the difference between the center of 2D bounding box and the projected 3D centroid position $(h_{\textrm u}, h_{\textrm v}) \in \mathbb{R}^2$, together with the depth of the centroid $h_{\textrm dep}$. We initialize $h_{\textrm u} = h_{\textrm v} = 0$, whereas $h_{\textrm dep}$ are set $h_{\textrm dim}$ to the mean depth and dimensions. We also initialize $h_{\textrm rot}, h_{\textrm sh}, h_{\textrm tx}$ to random numbers sampled from a Gaussian with mean $\mu=0$ and standard deviation $\sigma=0.1$.

After this initialization, we run an iterative pipeline of 1) rendering, 2) projective loss computation over multiple different cues and 3) backpropagation to our latent object variables. To produce a rendering, we feed $h_{\textrm sh}$ and $h_{\textrm tx}$ through generative models to produce object shape and texture. We then rescale and rotate using $h_{\textrm dim}$ and $h_{\textrm rot}$, place the object in the 3D space using $h_{\textrm loc}$ and finally render an RGB/D image pair with the differentiable renderer implementation from~\cite{kato2018neural}. Note that we work at metric scale and therefore require known camera intrinsics during optimization.

%We mask out the object regions in the input image using instance segmentation and use it as a background of the rendered image. 

%To fully utilize the information that can be obtained from input images, we leverage pre-trained models for all involved tasks. Concretely, we use pre-trained 2D detection, instance segmentation, monocular depth estimation, and object generation models. 

%We assume that intrinsic and extrinsic parameters of a camera are known during inference, as with~\cite{manhardt2019roi,simonelli2019disentangling,simonelli2019single}.

%%%%%%%% figure %%%%%%%%%%%%%%%%%%%%%%%%

%%%%%%%% figure %%%%%%%%%%%%%%%%%%%%%%%%
\begin{figure*}[t]
    \begin{center}
        \includegraphics[scale=0.75]{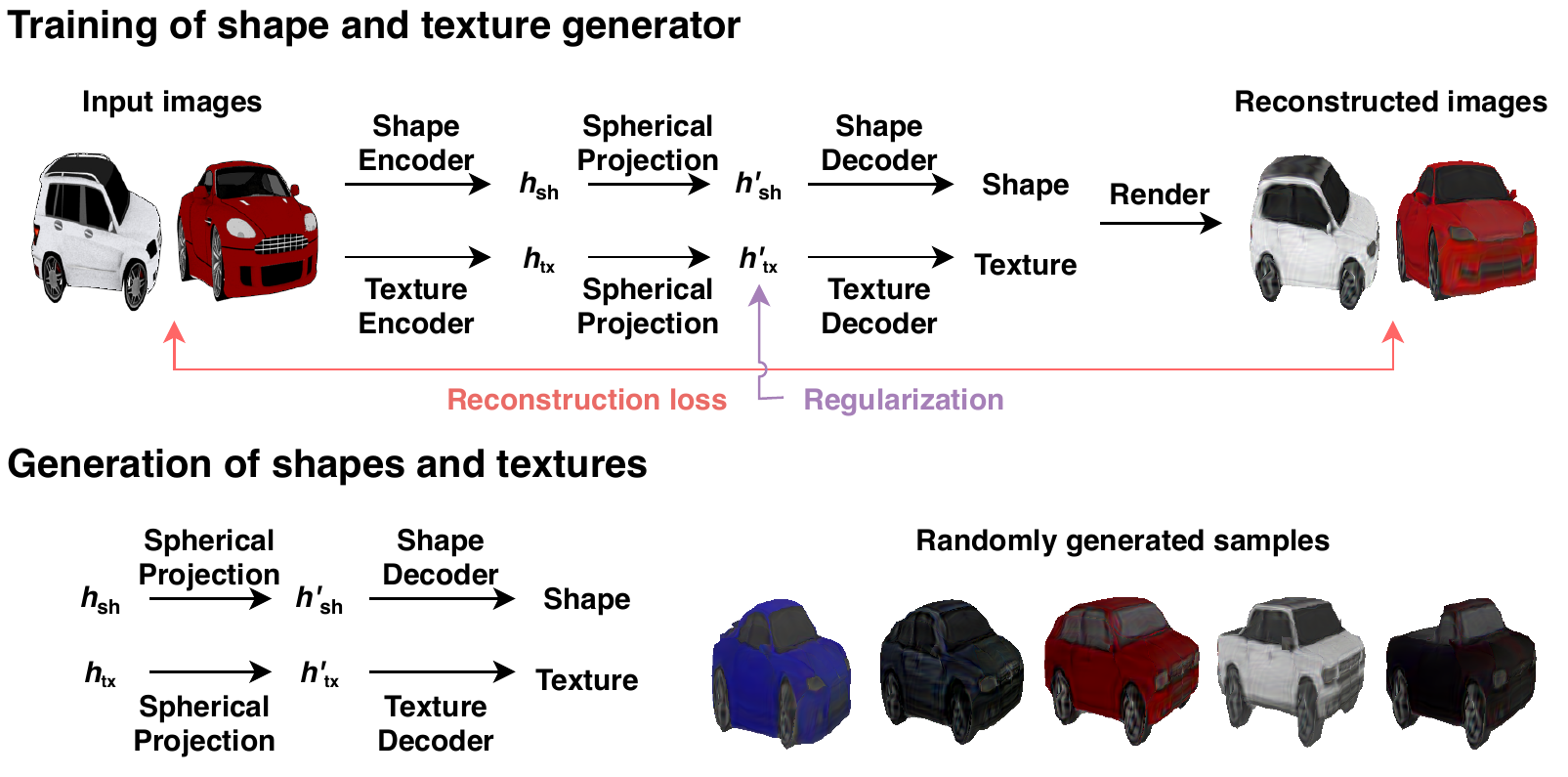}
    \end{center}
    \caption{Illustration of our shape and texture generator. During training, images of cars are encoded to shape and texture vectors that live on hyperspheres. These vectors can then be decoded to 3D shapes and texture maps, which in turn are rendered and used to compute reconstruction losses. After training, the decoder part can be used independently to generate novel textured shapes. }
    \label{fig:car_generation}
\end{figure*}

%\subsection{Requirements for Optimization}

\subsubsection{2D Object Detection and Segmentation}
To identify all instances in the scene, we use an off-the-shelf Mask R-CNN~\cite{he2017mask} model with the X-152 backbone from detectron2~\cite{wu2019detectron2}, trained on COCO \cite{tsung2014mscoco} and thresholded at 0.1.

To compute our rendering losses, we need a clear foreground/background separation. In order to produce a background mask, we compute a union over all detection masks to first obtain a foreground mask, and then invert it. We will evaluate the majority of our experiments with this approach. Alternatively, one can also leverage semantic segmentation for the separation and we present an ablative comparison in the experimental section.     

\subsubsection{Rendering}
We use the publicly available differentiable renderer from Kato et al.~\cite{kato2018neural} and although they introduced their method for color and silhouettes only, we extend it for rendering depth maps in a differentiable way. We exploit the simple fact that a depth map is differentiable with respect to the z-coordinates of a surface without any approximations because the depth value at a pixel is computed by a weighted sum of z-coordinates of the surfaces at the pixel.

\subsubsection{Shape and Texture Generation}
Differentiable rasterization allows us to establish losses between renderings and an input image ({\it render-and-compare}). This approach was initially adopted by 3D-RCNN~\cite{kundu2018cvpr} in which shapes were projected to 2D silhouettes. Unfortunately, silhouettes retain only a fraction of the original information and have an especially strong impact on rotation estimation. To leverage image information as much as possible, we therefore additionally reconstruct textures and render a joint representation of textured shapes. 

To encode a variety of object shapes and appearances, we employ a generative model that is able to produce a mesh and a texture map for given inputs. Concretely, we use the single-image 3D object reconstruction method presented in~\cite{kato2019learning} and provide an illustration in Figure \ref{fig:car_generation}. This method adopts an encoder-decoder architecture that transforms an image into a low-dimensional latent space which in turn can be used to generate a 3D object. Even though we train the entire encoder-decoder architecture beforehand, we discard the encoder and only use the decoder later on. 

We would like our generator to produce plausible shapes and textures for any  decoding input $h_{\textrm sh}$ and $h_{\textrm tx}$, ensuring that our latent spaces are smoothly traversable without collapsing. Therefore, unlike ~\cite{kato2019learning}, we project the hidden vectors onto a unit hypersphere, similar to~\cite{zakharov2020autolabeling}. To distribute $h'_{\textrm sh}$ and $h'_{\textrm tx}$ uniformly on the hypersphere, we employ a regularization technique where we take random samples on the sphere and pull the closest latent vectors towards those random samples. Let $r_{\textrm sh}$ and $r_{\textrm tx}$ be random samples uniformly distributed  on the hyperspheres of $D_{\textrm sh}$ and $D_{\textrm tx}$ dimensions. We add the following term to the loss function:
\begin{eqnarray}
    L_h = \frac{1}{N_b} \left( \sum_{i=1}^{N_b} \min_{j} |r^i_{\textrm sh} - h^j_{\textrm sh}|_1 + \sum_{i=1}^{N_b} \min_{j} |r^i_{\textrm tx} - h^j_{\textrm tx}|_1 \right).
\end{eqnarray}
$N_b$ is the size of minibatch, and $h'^i_{\textrm sh},h'^i_{\textrm tx}, r^i_{\textrm sh},r^i_{\textrm sh}$ represent $i$-th sample in the minibatch. The encoder-decoder architecture is trained using synthetic views of 3D car models taken from the ShapeNet dataset~\cite{chang2015shapenet}. We found a latent dimensionality of $8$ for the shape $D_{\textrm sh}$ and texture $D_{\textrm tx}$ spaces to work quite well. 

\subsubsection{Monocular Depth Estimation}
To weaken the projective entanglement between object dimensions, locations and 2D appearance, we employ depth maps extracted by a pre-trained, state-of-the-art monocular depth estimator that is trained with self-supervision (PackNet~\cite{guizilini20193d}). The scale of the estimated depth is calibrated from weak supervision in the form of ego-camera velocity.

%making an accurate prediction only from 2D space, without using any 3D information like LiDAR or supervision, is an ill-posed problem. To resolve this ambiguity, we use monocular self-supervised depth estimation methods that are trained without LiDAR or 3D supervision. Though the scale of the estimated depth is not calibrated in most of the methods, the real-scale depth can be obtained without accessing LiDAR data by using other weak supervision such as the velocity of an ego camera. We leverage the depth maps extracted by a pre-trained, a state-of-the-art monocular depth estimator that is trained with self-supervision. (e.g., PackNet~\cite{guizilini20193d}).

\subsection{Loss Functions}
\label{sec:method:loss}
For self-supervision we propose four loss functions between renderings and image evidence: $L_p$ which penalizes pixel color deviations, $L_b$ to maximize 2D bounding box overlap, $L_m$ for silhouette alignment,and $L_d$ for depth map differences.
We notate an input image of $H \times W$ pixels as $I^{\textrm RGB} \in \mathbb{R}^{H \times W \times 3}$, the foreground map and depth map estimated from the image as $I^{\textrm F} \in \mathbb{R}^{H \times W}$ and $I^{\textrm D} \in \mathbb{R}^{H \times W}$, respectively. In addition, assuming that the number of detected objects is $N_o$, we notate the cropped and resized region of the $i$-th object from these maps as $I^{\textrm RGB}_i \in \mathbb{R}^{H_c \times W_c \times 3}$, $I^{\textrm F}_i \in \mathbb{R}^{H_c \times W_c}$, and $I^{\textrm D}_i \in \mathbb{R}^{H_c \times W_c}$. For the reconstructed counterparts, we use the same notations with {\it hat}.

Additionally, we use a regularization term $L_{\textrm dim}$ to penalize unrealistic dimensions. We define it as the L1 distance between the estimated dimension and the mean dimension of objects in the dataset. The overall loss function is defined as the weighted sum of all the loss functions and the regularization term:
\begin{equation}
    L = \lambda_p L_p + \lambda_b L_b + \lambda_m L_m + \lambda_d L_d + \lambda_{\textrm dim} L_{\textrm dim}
    \label{eq:total-loss}.
\end{equation}

\subsubsection{Pixel Map Difference}
The pixel map difference loss is defined as L1 distance between rendered image and input image by using equation 
\begin{equation}
    L_p = \frac{1}{H_c W_c} \sum_{i=1}^{N_b} \sum_{j=1}^{H_c} \sum_{k=1}^{W_c}|I^{\textrm RGB}_{ijk} - {\hat I}^{\textrm RGB}_{ijk}|.
\end{equation}

\subsubsection{Bounding Box Difference}
We compute the bounding boxes of the objects from the rendered image. Let $\textrm{IoU}^B_i$ be the intersection over union between input and rendered 2D bounding boxes of $i$-th object. We define the bounding box loss as 
\begin{equation}
    L_b = \sum_{i=1}^{N_b} \textrm{maximum}(1 - \textrm{IoU}^B_i - t_b, 0).
\end{equation}
As the predicted bounding boxes may be noisy, we allow the IoU to have a small error margin of up to $t_b$. This loss is differentiable with respect to the vertices of the reconstructed shapes as the bounding box is computed by differentiable projection of vertices to image coordinates. Fitting estimated 3D bounding boxes to detected 2D bounding boxes is a well-known approach~\cite{mousavian20173d}, and some works~\cite{min2019multi} use IoU as a metric. Different from these works, we compute projected bounding boxes using 3D shapes instead of 3D bounding boxes.

\subsubsection{Silhouette Difference}
Concretely, assuming that $I^{\textrm F}_{ijk} = 1$ if the pixel $jk$ of the $i$-th image is foreground and $I^{\textrm F}_{ijk} = 0$ if background, the IoU of $i$-th image is

\begin{eqnarray}
    \textrm{IoU}^F_i =
    \frac{
        \sum_{j=1}^{H_c} \sum_{k=1}^{W_c} 
            I^{\textrm F}_{ijk} {\hat I}^{\textrm F}_{ijk}
    }{
        \sum_{j=1}^{H_c} \sum_{k=1}^{W_c} 
            I^{\textrm F}_{ijk} + {\hat I}^{\textrm F}_{ijk} - 
            I^{\textrm F}_{ijk} {\hat I}^{\textrm F}_{ijk}
    }, \\
    L_m = \sum_{i=1}^{N_o}
    L_{m_{i}} = \sum_{i=1}^{N_o} (1 - \textrm{IoU}^F_i).
\end{eqnarray}

\subsubsection{Depth Map Difference}
The depth map loss is defined as the L1 difference between the rendered depth and model-predicted depth by using the formula
\begin{equation}
    L_d = \frac{1}{H_c W_c} \sum_{i=1}^{N_b} \sum_{j=1}^{H_c} \sum_{k=1}^{W_c}  I^{\textrm F}_{ijk} {\hat I}^{\textrm F}_{ijk} |I^{\textrm D}_{ijk} - {\hat I}^{\textrm D}_{ijk}|.
\end{equation}

Note that during the optimization, all the pre-trained networks are used with fixed weights and are not optimized. We disable back-propagation for ${\hat I}^{\textrm F}_{ijk}$ and back-propagate gradients only through ${\hat I}^{\textrm D}_{ijk}$.

\subsection{Escaping Rotational Local Minima}
\label{sub_sec:local_minima}
Estimating rotation with render-and-compare approaches can easily lead to local minima~\cite{insafutdinov2018unsupervised,tulsiani2018multi} due to non-linear objectives and visual ambiguities. To deal with this issue, we explore several rotations at every step of the optimization. We sample 4 different variations ($h_{\textrm rot}$, $h_{\textrm rot} \pm 15\degree$, -$h_{\textrm rot}$) as well as a random angle sampled from a uniform distribution. If an angle $r_0$ gives a lower error with the perceptual metric~\cite{zhang2018unreasonable} than the one used at current iteration, we set $h_{\textrm rot}$ to $r_0$. Also, the size of the input and the reconstructed images are adjusted to ensure that the area of the object bounding boxes is the same, and the background region is masked out using the estimated segmentation mask.

\subsection{Detection Confidence Score}
MonoDIS~\cite{simonelli2019disentangling} proposes to learn confidences of 3D detections by weighting confidences of corresponding 2D detections. As their method requires ground-truth 3D bounding boxes for confidence training, it is not applicable in a straight-forward manner to our problem. Instead, we propose to weight 2D confidences using our reconstruction errors after optimization. To this end, we compute the silhouette reconstruction loss $L_{{m}_{i}}$ of the $i$-th image, and the ratio of protrusion $b_i$ between rendered bounding box and detection bounding box. Our confidence score $c_{\textrm 3D}$ is then defined by using the formula 
\begin{equation}
    c_{\textrm 3D} = c_{\textrm 2D} e^{-\alpha_m L_{{m}_{i}}} e^{-\alpha_b b_{i}},
\end{equation}
with two hyperparameters $\alpha_m, \alpha_b$ and the original 2D detection confidence $c_{\textrm 2D}$.

%% file: 05_experiments.tex
\section{Experiments}
\label{sec:experiments}

We evaluate the proposed method on KITTI 3D~\cite{geiger2012we}, one of the most popular benchmarks for 3D object detection in autonomous driving. KITTI 3D is composed of synchronized RGB images/LiDAR frames, for which annotations of 2D bounding boxes, 3D object location, 3D object dimension and 1D object rotation angle along the {\it y}-axis (yaw) are provided. Unlike other approaches, our method does not require 3D ground-truth or LiDAR observations in any part of the pipeline. For a fair comparison to similar work, we use the same training and validation splits proposed in~\cite{chen20153d} and focus on the {\it car} category.

\subsubsection{Evaluation Metrics}
\label{subsec:exp:metrics}
KITTI 3D uses different AP metrics to measure 2D and 3D detection accuracy at different cut-off thresholds (usually 0.5 and 0.7). For each detection, the IoU overlap with a ground truth is computed either on the image plane (2D), in bird's eye view (BEV), or in volumetric 3D space.   
Following a recent suggestion by~\cite{simonelli2019disentangling}, we use the $AP_{|R_{40}}$ metric that is currently used in the official KITTI 3D benchmark leaderboard, instead of the older $AP_{|R_{11}}$ metric. For reference and a fair comparison to previously established work, we also evaluate against the $AP_{|R_{11}}$ metric and share the results in the supplement.

\subsubsection{System Configuration}
\label{subsec:exp:sysconfig}

All experiments are performed on a Linux-based cluster with 8 V100 (32GB) GPUs. Running the full pipeline on the KITTI 3D {\it validation} dataset takes approximately one day.

\subsubsection{Hyperparameters Tuning}
\label{subsec:exp:hyperparameters}

 The hyperparameter values are set to $\lambda_b = \lambda_m = \lambda_p = 1$, $\lambda_d = 0.2$, $\lambda_{dim} = 0.1$, and $t_b = 0.1$. We use the Adam optimizer ~\cite{kingma2014adam} with $\alpha=0.03$, $\beta_1=0.5$ and $\beta_2=0.9$. For each detected object, we run the optimization pipeline for $150$ iterations. The hyperparameters for confidence weighting are set to $\alpha_m = 0.1$ and $\alpha_b = 0.03$.

\subsection{Comparison to SoTA}
\label{subsec:exp:benchmark}
As a first experiment, we run our pipeline on the validation set and show our quantitative results in Table~\ref{table:val_ap40} where we compare our inferred 3D boxes against the ground truth labels. As can be seen, we position ourselves between the two leading monocular 3D detectors MonoDIS~\cite{simonelli2019disentangling} and MoVI-3D~\cite{simonelli2019single} for the easy instances and fall slightly behind for the moderate and difficult cases. This shows that our self-supervised 3D estimation method can be competitive with fully supervised detectors trained on manually labeled 3D ground truth cuboids. For more difficult instances, the monocular depth maps become noisier at longer ranges, there is an increasing amount of occlusion, and there are smaller objects, all factors that challenge our losses for bounding box, silhouette, and pixel differences.
The autolabeling solution from \cite{zakharov2020autolabeling} leverages shape priors similar to us, yet has access to LiDAR observations that greatly benefit their scale and 3D location estimation. Nonetheless, especially for the more challenging 3D AP metric, we can retrieve much more accurate object estimates overall. One of our major differences over LiDAR equipped SDFLabel paper is the scale regression method. SDFLabel often fail the strict 3D AP metric due to small deviations in 3D scale regression, especially at range where very few LiDAR points are on the object. On the other hand, we leverage the dimension statistics of the cars as a strong additional prior over generic driving scene instances. While SDFLabel regress directly on metric scale, we regressed the deviation from the statistics.

%%%%%%%% table %%%%%%%%%%%%%%%%%%%%%%%%
% Comparison with SOTA, AP40, validation set
\begin{table}[b]
    \small
    \begin{center}
        \begin{tabular}{l|c|c|ccc|ccc}
            \toprule
            \multirow{2}{*}{Method} & \multirow{2}{*}{Supervised} & \multirow{2}{*}{LiDAR} & \multicolumn{3}{c|}{3D detection} & \multicolumn{3}{c}{Bird’s eye view} \\ 
             ~ & ~ & ~ & Easy & Moderate & Hard  & Easy & Moderate & Hard \\ 
            \midrule   
            MonoDIS~\cite{simonelli2019disentangling} & \checkmark & & 11.06 & 7.60 & 6.37 & 18.45 & 12.58 & 10.66  \\
            MoVI-3D~\cite{simonelli2019single} & \checkmark & & 14.28 & 11.13 & 9.68 & 22.36 & 17.87 & 15.73   \\
            SDFLabel~\cite{zakharov2020autolabeling} & & \checkmark & 1.23 &  0.54 & n/a & 15.7 & 10.52 & n/a \\   
            \midrule
            MonoDR (ours) & & & 12.50 &  7.34 & 4.98 & 19.49 & 11.51 & 8.72 \\
            \bottomrule
        \end{tabular}
    \end{center}
    \caption{Evaluation of different monocular 3D detection methods: We report $AP_{|R_{40}}$ on the KITTI 3D {\it validation} set. The values are calculated assuming an intersection-over-union (IoU) between the predicted and ground-truth bounding boxes (in bird's-eye view) of at least 0.7.}
    \label{table:val_ap40}
\end{table}
%%%%%%%% table %%%%%%%%%%%%%%%%%%%%%%%%

%We pick three images from the dataset and visualize the monocular 3D detection results produced by our method in . The images contain different numbers of objects, which are located at different distances and vary in rotation. Estimating the 3D location from a monocular image is difficult especially for far objects, so a depth map is used to recover the correct one.

\begin{figure*}[h!]
    \begin{center}
        \includegraphics[width=1.0\linewidth]{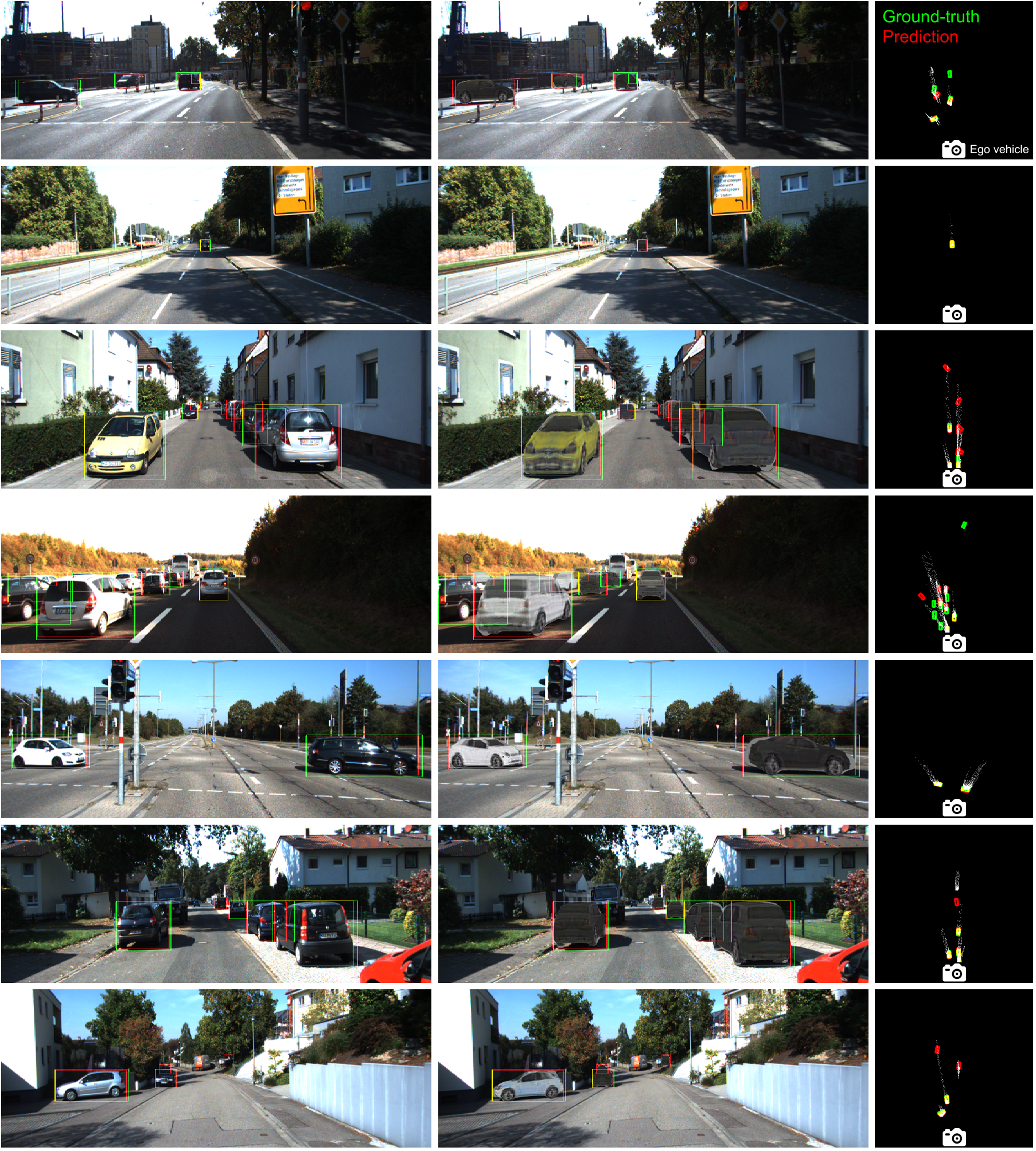}
    \end{center}
    \caption{Representative results obtained through our method: Raw images with predicted 2D bounding boxes and instance masks are shown in the left column. Reconstructed images (with the projection of an estimated 3D shape onto the image plane) are shown in the middle column. Ground-truth (green)/predicted (red) 3D bounding boxes in the bird's-eye view (BEV) are shown in the right column. The white points in BEV represent projections of object point clouds generated from predicted depth maps. The ground truth information is only used for visualization purposes. }
    \label{fig:result1}
\end{figure*}

We also present some qualitative results in Figure~\ref{fig:result1}. It can clearly be seen that our priors provide enough guidance to estimate the correct 3D locations and rotations of the objects even if the predicted depth map is noisy. The depth map provides weak supervision for the 2D loss to estimate the correct object scale and rough 3D location but is ultimately regularized by shape and texture. It is also evident that optimizing for the 2D bounding box IoU loss $L_{b}$ results in 3D shape projections that tightly fit into the predicted 2D bounding boxes. Besides, the  regularization loss $L_{dim}$ on dimensions prevents the predicted shapes from being too different from the mean object shape. %Hence, the entanglement between dimension, location and 2D projection can be resolved.

%Table~\ref{table:val_ap40} shows the evaluation results of different monocular 3D detection methods on the KITTI 3D {\it validation} set. For a fair comparison with other papers, we also provide the $AP_{|R_{11}}$ results in the Appendix. The accuracies of existing methods that do not use LiDAR data for training and inference \cite{simonelli2019disentangling,simonelli2019single} are also shown. Our method is comparable to the state-of-the-art~\cite{simonelli2019single}, which requires supervision. Even though the results indicate that supervised 3D detection is still a challenging problem, it can be solved through simple fitting-based methods. This is in part thanks to recent advancements in 2D object detection, instance segmentation and self-supervised monocular depth estimation.

\subsection{Ablation Studies}
\label{subsec:exp:ablation}

\subsubsection{Reconstruction Loss}
%%%%%%%% table %%%%%%%%%%%%%%%%%%%%%%%%
% "Which 2D loss is important?"
% "Is depth needed?" 
% "What is a good way to integrate depth?"
\begin{table*}[h!]
    \small
    \begin{center}
        \begin{tabular}{l|ccc|ccc}
            \toprule
            \multirow{2}{*}{Reconstruction loss} & \multicolumn{3}{c|}{3D Detection} & \multicolumn{3}{c}{Bird’s eye view} \\
             ~ & Easy & Moderate & Hard  & Easy & Moderate & Hard \\ 
            \midrule
            full ($L$) & 12.50 &  7.34 & 4.98 & 19.49 & 11.51 & 8.72 \\
            % w/o bounding box ($L - \lambda_b L_b$) & 10.57 & 6.04 & 4.31 & 17.67 & 10.08 & 7.54 \\
            % w/o silhouette ($L - \lambda_m L_m$) & 7.19 & 4.17 & 3.03 & 13.29 & 8.11 & 5.97 \\
            % w/o pixel ($L - \lambda_p L_p$) & 11.60 & 6.32 & 4.46 & 17.64 & 10.02 & 7.43 \\
            w/o depth ($L - \lambda_d L_d$) & 4.82 & 2.88 & 1.91 & 7.96 & 4.96 & 3.64 \\
            depth only ($L_d$) & 0.00 & 1.00 & 0.00 & 0.00 & 0.00 & 0.00 \\
            \bottomrule
        \end{tabular}
    \end{center}
    \caption{Evaluation of different reconstruction losses on KITTI 3D {\it val} set.}
    \label{table:2d_loss_breakdown}
\end{table*}
%%%%%%%% table %%%%%%%%%%%%%%%%%%%%%%%%

 %To this end the bounding box difference loss $L_b$, the silhouette difference loss $L_m$, the pixel map difference loss $L_p$ and the depth map difference loss $L_d$, to $L$. We evaluate on the different subsets (Easy, Moderate, Hard) of the KITTI 3D dataset and report the $AP_{|R_{40}}$ values.

Firstly, we analyze the contribution of each individual loss component on the final outcome. Table \ref{table:2d_loss_breakdown} shows that all losses have a significant impact. Among them, $L_d$ is the most important, as the accuracy sharply drops by over 60\% when it is removed from the loss function. This shows that using 2D loss functions alone is not as effective, and that even noisy knowledge of scene depth (either implicitly or explicitly) is essential for monocular object estimation tasks. On the other hand, optimizing only for $L_d$ leads to an accuracy close to zero. This signifies that using depth alone is apparently not enough for accurate reconstruction. In such cases, the optimization is disregarding object boundaries and physical extents. For example, the detected and reconstructed silhouettes should overlap, but the loss function does not enforce this constraint. Additionally, the objects can warp into any possible size. 

We experimented to integrate depth information by backprojecting depth maps into 3D point clouds similar to Pseudo-LiDAR~\cite{you2019pseudo}. We optimized the Hausdorff distance between predicted meshes and the point clouds, and also experimented with Procrustes variants to support rotation, translation and scale. However, both results were very similar to our current {\it depth-only} results. 

Another strong impact can be seen from the silhouette loss $L_m$. This loss back-propagates 2D silhouette mismatches to penalize the reconstruction of the 3D shape. Along with the regularization loss of dimensions $L_{dim}$, it helps converging towards a realistic shape that matches the predicted mask. 

On the other hand, the impact of bounding box IoU loss $L_b$ and pixel color loss $L_p$ have less impact on the final accuracy. $L_b$ strongly enforces 2D projection of the reconstruction to be within the bounding box limits, which can be interpreted as an auxiliary support for the silhouette loss $L_m$. Similarly, the pixel loss impacts primarily the texturing of objects and helps mostly for minor textural misalignments but can hardly recover larger displacements or affect the shape. 

\subsubsection{Confidence Weighting}

%%%%%%%% table %%%%%%%%%%%%%%%%%%%%%%%%

%c_{\textrm 3D} = c_{\textrm 2D} e^{-\alpha_m L_{m}_{i}} e^{-\alpha_b b_{i}}
% "Is confidence weighting by reconstruction good?"
\begin{table*}[t]
    \small
    \begin{center}
        \begin{tabular}{l|ccc|ccc}
            \toprule
            \multirow{2}{*}{Confidence weighting} & \multicolumn{3}{c|}{3D detection} & \multicolumn{3}{c}{Bird’s eye view} \\
             ~ & Easy & Moderate & Hard  & Easy & Moderate & Hard \\ 
            \midrule
            full ($c_{3D}$) & 12.50 &  7.34 & 4.98 & 19.49 & 11.51 & 8.72 \\
            w/o silhouette recon. ($c_{2D} e^{-\alpha_b b_{i}}$) & 11.19 & 6.81 & 4.61 & 17.72 & 10.80 & 8.20 \\
            w/o box protrusion ($c_{2D} e^{-\alpha_m L_{{m}_{i}}}$) & 12.43 & 7.29 & 4.89 & 19.29 & 11.34 & 8.64 \\
            w/o both ($c_{2D}$) & 10.74 & 6.48 & 4.40 & 17.19 &10.36 & 7.95 \\
            \bottomrule
        \end{tabular}
    \end{center}
    \caption{Evaluation of different confidence weighting schemes on the {\it val} set.}
    \label{table:confidence_weighting}
\end{table*}
%%%%%%%% table %%%%%%%%%%%%%%%%%%%%%%%%

In Table \ref{table:confidence_weighting}, we illustrate the impact of our proposed confidence weighting schemes. In the last row, we present the accuracies calculated by using the scores obtained from 2D bounding box detection without applying any confidence weighting. In the first row, we present the result after applying the confidence weighting over all scores. Applying both weights improves the accuracy by 11\% - 16\%. The second row shows that weighting the scores by using the silhouette reconstruction loss value significantly impacts the increase. 

\subsubsection{Effect of the Segmentation Masks}

%%%%%%%% table %%%%%%%%%%%%%%%%%%%%%%%%
% "What is the impact of 2D bounding box detectors on accuracy"
\begin{table*}[h!]
    \small
    \begin{center}
        \resizebox{\textwidth}{!}{\begin{tabular}{c|cc|ccc|ccc|ccc}
            \toprule
            \multirow{2}{*}{Model} & \multicolumn{1}{c}{Box AP} & \multicolumn{1}{c|}{Mask AP} & \multicolumn{3}{c|}{2D detection} & \multicolumn{3}{c|}{3D detection} & \multicolumn{3}{c}{Bird's Eye View} \\ 
             ~ & COCO & COCO & Easy & Moderate & Hard  & Easy & Moderate & Hard & Easy & Moderate & Hard   \\ 
            \midrule   
            Mask R-CNN X152   & 50.2 & 44.0 & 82.24 & 74.82 & 59.34  & 12.50 & 7.34 & 4.98 & 19.49 & 11.51 & 8.72 \\
            Panoptic R101-FPN & 42.4 & - & 78.59 & 72.17 & 59.66  & 9.26 & 5.84 & 4.35 & 15.42 & 9.97 & 7.44 \\
            \bottomrule
        \end{tabular}
        }
    \end{center}
    \caption{Different segmentation networks, on KITTI 3D {\it validation} set.}
    \label{table:detection_networks}
\end{table*}
%%%%%%%% table %%%%%%%%%%%%%%%%%%%%%%%%

In Table \ref{table:detection_networks}, we present the impact of the chosen 2D object detection and segmentation network on the final accuracy. We compared the results of Mask R-CNN X-152 \cite{he2017mask} and Panoptic Segmentation R101-FPN, both taken from detectron2\cite{wu2019detectron2} and pretrained on COCO \cite{tsung2014mscoco}. Mask R-CNN has significantly better mAP than the panoptic segmentation model, especially for the easy and moderate instances. Even though panoptic segmentation provides better semantic masks, it provides worse 2D detection performance for occluded objects. This is due the general nature of panoptic segmentation, since it derives its bounding boxes tightly around the segmentations. As our method is strongly dependent on the tightness of 2D bounding boxes, panoptic segmentation has an overall negative impact on the accuracy.

\subsection{Limitations and Failure Cases}
\label{subsec:exp:limitations}

\begin{figure*}[h!]   
    \centering
    \subfloat[Example of failures caused by occlusion. The cars on the right side (top row) and the car on the left side (bottom row) are partially visible and the estimation of their 2D bounding boxes is not accurate, which leads to large errors in rotation estimation.]{
        \includegraphics[width=1.0\linewidth]{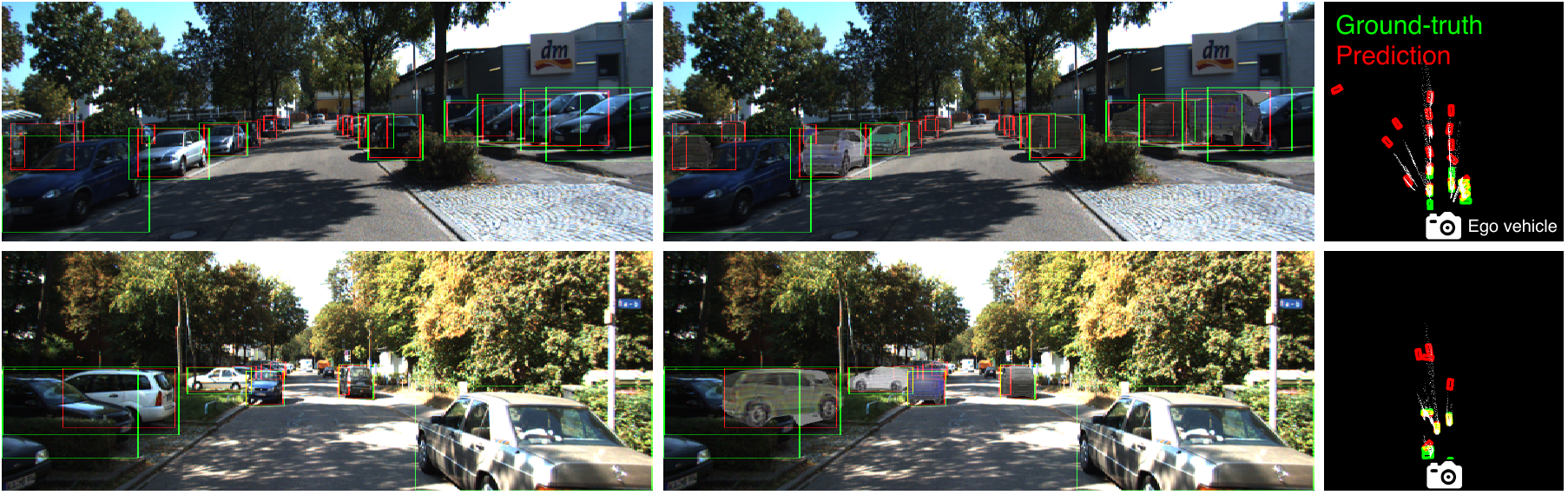}
        \label{subfig:limitations:occlusion}
    }
    \hfill
    \subfloat[Example of a failure caused by a non-tight 2D bounding box. The inaccurate 2D bounding box estimation leads to a large error in 3D shape estimation and pose translation.]{
        \includegraphics[width=1.0\linewidth]{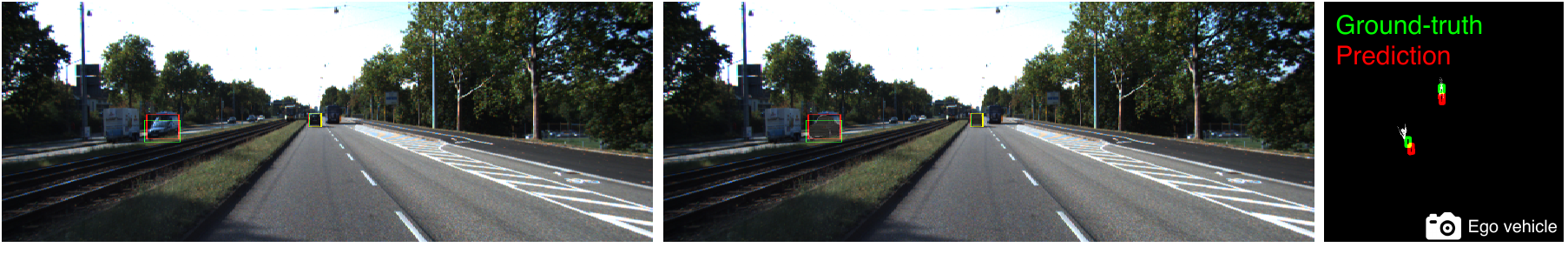}
        \label{subfig:limitations:tightness}
    }
    \caption{Two examples of cases where our method fails to estimate accurate object pose translations or rotations.  Ground-truth (green)/predicted(red) 3D bounding boxes in the bird’s-eye view (BEV) are shown in the right column. The white points in BEV represent projections of object point clouds generated from predicted depth maps. The ground truth information is only used for visualization purposes.}
    \label{fig:limitations}
\end{figure*}

The quality of the estimated bounding boxes and segmentation masks clearly impacts the rendering used for computing the loss. Inaccuracies in their estimation will be reflected in the quality of the estimated 3D shape and object pose.
Figure~\ref{subfig:limitations:occlusion} shows two example of failure cases due to stronger car occlusions. Because the 2D bounding boxes used in this work are not {\it amodal}, they only confine the visible parts of the cars. In this case, optimization is done by projecting the 3D shape onto a partial view of the car, which leads to a large error in the rotation estimation. This problem could be solved by fine-tuning the detector on a dataset with annotated amodal 2D bounding boxes.

%%%%%%%% figure %%%%%%%%%%%%%%%%%%%%%%%%

%%%%%%%% figure %%%%%%%%%%%%%%%%%%%%%%%%

The size of the estimated 3D shape is also constrained by the size of the estimated 2D bounding box. Figure~\ref{subfig:limitations:tightness} shows an example of a failure case where the 2D bounding box is larger than the object (non-tight). Because the optimization requires the projection of the 3D shape to fit the 2D bounding box, the shape is constrained to be larger than it should be. In this case, a large translation error can be seen due to a shift in the center of mass.

The computational cost is currently a bottleneck of our method. The current computation time is proportional to the number of detected objects and the mean computation time per image on the KITTI {\it validation} set is three minutes. The major time is spent on escaping the local minima of the rotation \ref{sub_sec:local_minima} as we had to render multiple times per image.

%As a unsupervised method, use ours as auto labeling and train a faster detector using estimated labels like~\cite{TODO} would be a practical solution.

%% file: 06_conclusion.tex
\section{Conclusion}

We presented a self-supervised approach, based on differentiable rendering, for 3D shape reconstruction and localization of rigid objects from monocular images. Although used in the context of autonomous driving, our method is generally applicable to many categories and scenarios. We showed that it is possible to use noisy monocular depth and differentiable rendering in conjunction with learned object priors as an alternative to expensive 3D ground-truth labels or LiDAR information.

Future work should investigate alternative approaches to estimating depth from monocular images and registering point clouds. Although significantly improved over the last years, accurate long-range depth estimation is still a challenging problem. This information is of vital importance in autonomous driving, where the system often needs to make quick decisions. Having accurate information about the surrounding environment, as early as possible, allows for better risk assessment, planning and control.

Accurate and fast point cloud registration, on the other hand, is necessary in our pipeline for achieving self-supervision. Many of the current techniques are either slow or require an initial guess for accurate registration. This makes them impractical for applications where real-time inference is required.

%% file: 07_appendix.tex
\section{Implementation Details}

\subsection{Parameterization and Initialization}

An object's properties are defined as follows:
\begin{itemize}
    \item Location: $h_{\textrm{loc}} \in \mathbb{R}^3$
    \item Dimensions: $h_{\textrm{dim}} \in \mathbb{R}^3$
    \item Rotation: $h_{\textrm{rot}} \in \mathbb{R}$
    \item Shape encoding: $h_{\textrm{sh}} \in \mathbb{R}^{D_{\textrm{sh}}}$
    \item Texture encoding: $h_{\textrm{tx}} \in \mathbb{R}^{D_{\textrm{tx}}}$
\end{itemize}
Let $b_\textrm{top}, b_\textrm{left}, b_\textrm{bottom}, b_\textrm{right}$ be the coordinates of a detected 2D bounding box. We define the following:
\begin{align}
    c_u &= \frac{b_\textrm{top} + b_\textrm{bottom}}{2}, \\
    c_v &= \frac{b_\textrm{left} + b_\textrm{right}}{2}, \\
    s_u &= \frac{b_\textrm{bottom} - b_\textrm{top}}{2}, \\
    s_v &= \frac{b_\textrm{right} - b_\textrm{left}}{2}, \\
    u &= c_u + s_u h_u, \\
    v &= c_v + s_v (h_v + 0.5), \\
    z &= \mu_z + h_z \sigma_z
\end{align}

The location of the object, $h_{\textrm{loc}}$, is then expressed as
\begin{equation}
    h_{\textrm{loc}} = \textrm{proj} (u, v, z)
\end{equation}

Instead of directly optimizing for $h_{\textrm{loc}}$, we optimize for $h_u$, $h_v$, and $h_d$. $\mu_z$ and $\sigma_z$ are the mean and standard deviation of the $z$-coordinates of objects from the KITTI dataset. $\textrm{proj}$ is the projection operator defined in~\cite{manhardt2019roi}. We initialize $h_u$, $h_v$, and $h_d$ to zero.

The dimensions of the object, $h_{\textrm{dim}}$, are represented by
\begin{align}
    h_{\textrm{dim}} &= \mu_d + h_{\textrm{dim}}' \sigma_d,
\end{align}

Instead of directly optimizing for $h_{\textrm{dim}}$, we optimize for $h_{\textrm{dim}}'$. $\mu_d$ and $\sigma_d$ are the mean and standard deviation of dimensions of objects from the KITTI dataset. We initialize $h_{\textrm{dim}}'$ to zero.

Following~\cite{simonelli2019single}, the rotation $h_{\textrm{rot}}$ is defined as 
\begin{align}
    h_{\textrm{rot}} &= \textrm{arctan2}(h_{\textrm{rot}}^{\textrm{sin}}, h_{\textrm{rot}}^{\textrm{cos}}).
\end{align}

We initialize $h_{\textrm{rot}}^{\textrm{sin}}$ and $h_{\textrm{rot}}^{\textrm{cos}}$ by sampling from a Gaussian distribution with zero mean and a standard deviation of $0.1$.

The shape and texture encoding, $h_{\textrm{sh}}$ and $h_{\textrm{tx}}$, are directly optimized. We initialize them by sampling from a Gaussian distribution with zero mean and a standard deviation of $0.01$.

\subsection{Filtering of 2D Bounding Boxes}

%%%%%%%% figure %%%%%%%%%%%%%%%%%%%%%%%%
\begin{figure*}[t]
    \begin{center}
        \includegraphics[scale=0.7]{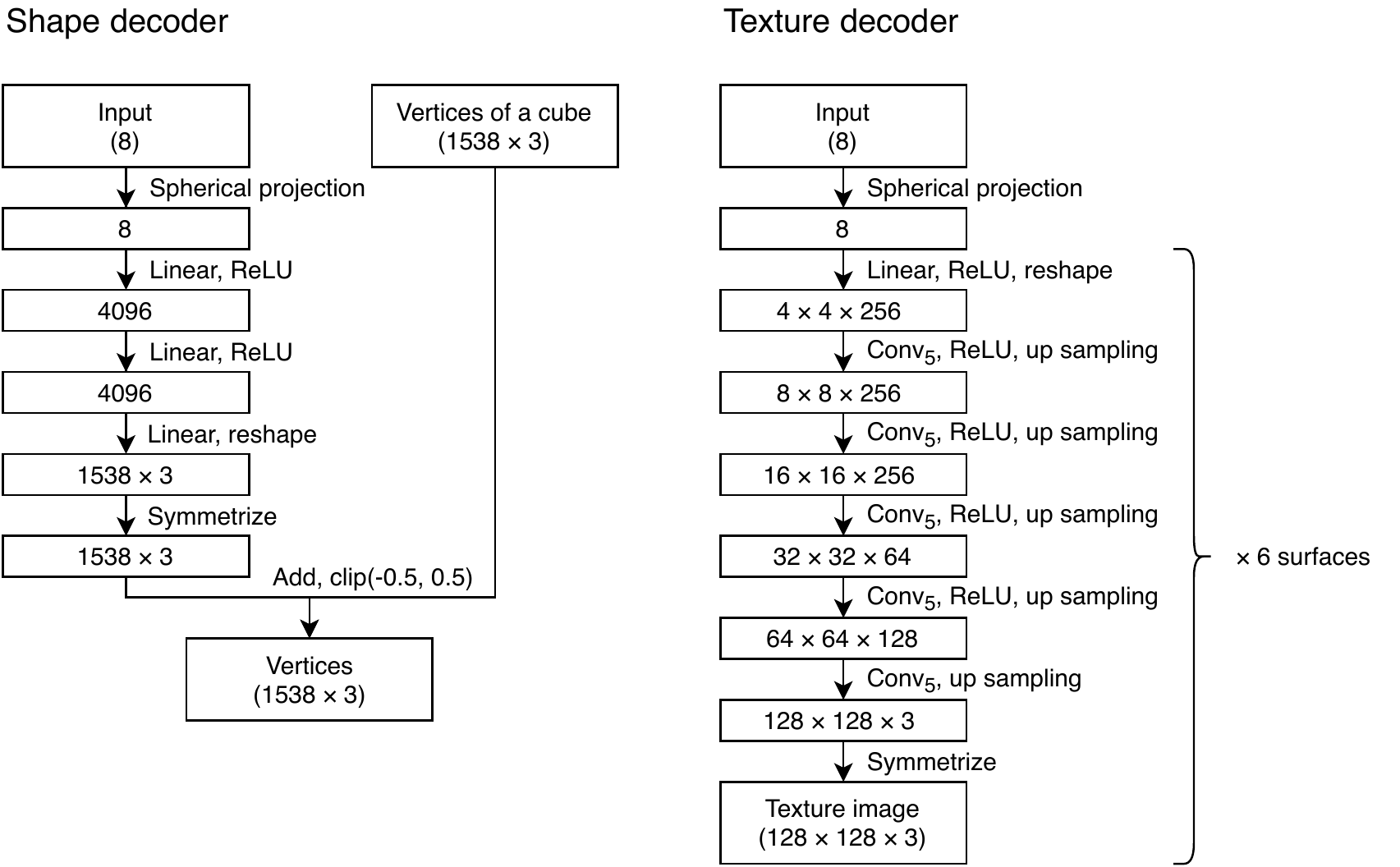}
    \end{center}
    \caption{We deform the vertices of a unit cube in order to generate the shape of an object. The cube is uniformly sampled to form a $17 \times 17 \times 17$ grid. The total number of vertices on the surfaces is, therefore, $17^3 - 15 ^ 3 = 1538$. Six texture images of $128 \times 128$ pixels are mapped onto the surfaces of the cube. $\textrm{Conv}_5$, in this figure, represents a 2D convolution operation that uses a $5 \times 5$ kernel.}
    \label{fig:car_generation_arch}
\end{figure*}
%%%%%%%% figure %%%%%%%%%%%%%%%%%%%%%%%%

We filter out 2D bounding boxes with a height smaller than 20 pixels because the minimum height of objects that are used for evaluation, on the KITTI dataset, is 25 pixels. We also remove 2D bounding boxes near the image boundary because the bounding box reconstruction loss is not meaningful if the bounding box is truncated.

\subsection{Network Architecture for Shape Generation}
We use the ResNet-18 architecture for encoding shape and texture. Fig.~\ref{fig:car_generation_arch} shows the network architecture used for decoding them.

\section{Additional Experimental Results}

Table~\ref{table:val_ap11} shows a comparison with other methods on the KITTI {\it validation} set, using the $AP_{|R_{11}}$ metric. Table~\ref{table:metrics_thresholds} shows a quantitative evaluation of our method on the KITTI {\it validation} set using different metrics and detection thresholds.

To confirm the effectiveness of shape reconstruction, we conducted two additional experiments. The first one uses randomly initialized object shapes and does not do any further shape optimization. The second one approximates object shapes with cuboids without shape optimization, as previously done in other works~\cite{mousavian20173d} that tackle the 3D detection problem.

%It is a well-known approach that measuring the correctness of 3D detection, assuming that object shapes are cuboids.

\begin{figure*}[h!]
    \centering
    \subfloat[3D detection with shape optimization.]{
        \includegraphics[width=0.32\linewidth,height=2cm]{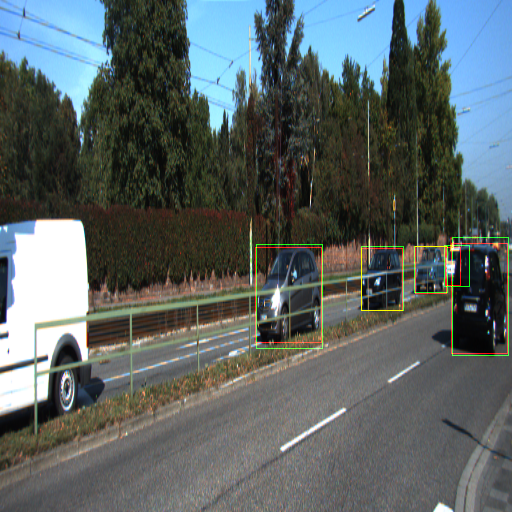}
        \includegraphics[width=0.32\linewidth,height=2cm]{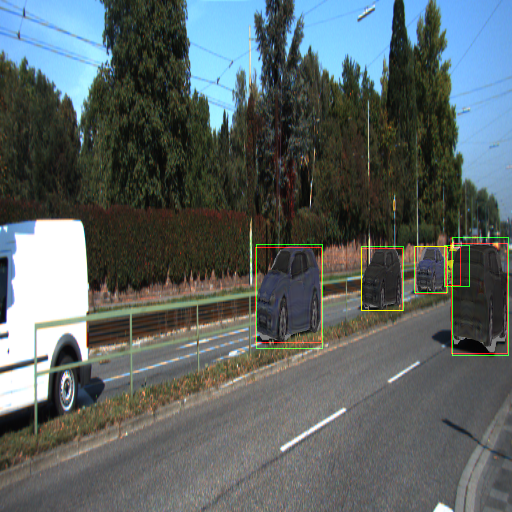}
        \includegraphics[width=0.32\linewidth,height=2cm]{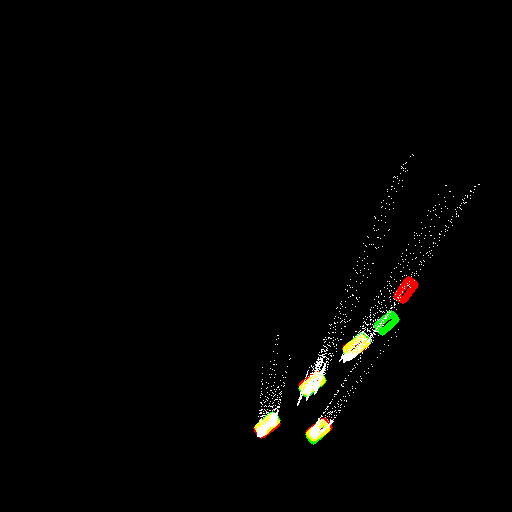}
    }
    \hfill
    \subfloat[3D detection without shape optimization (the shapes are randomly initialized).]{
        \includegraphics[width=0.32\linewidth,height=2cm]{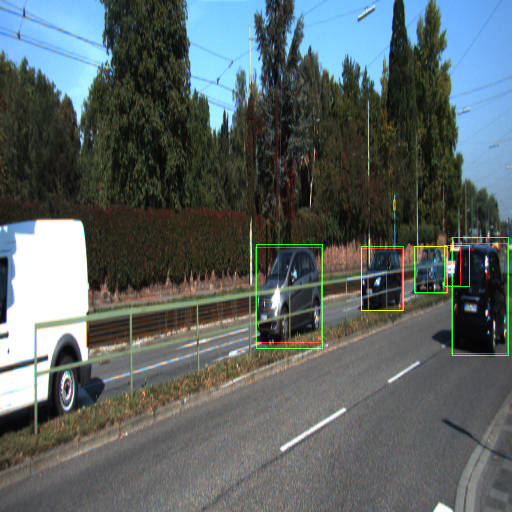}
        \includegraphics[width=0.32\linewidth,height=2cm]{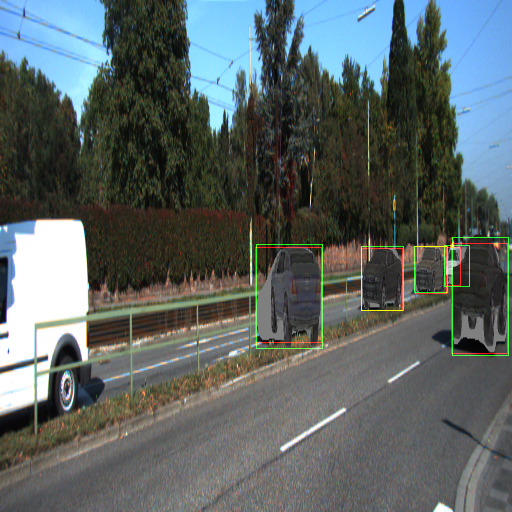}
        \includegraphics[width=0.32\linewidth,height=2cm]{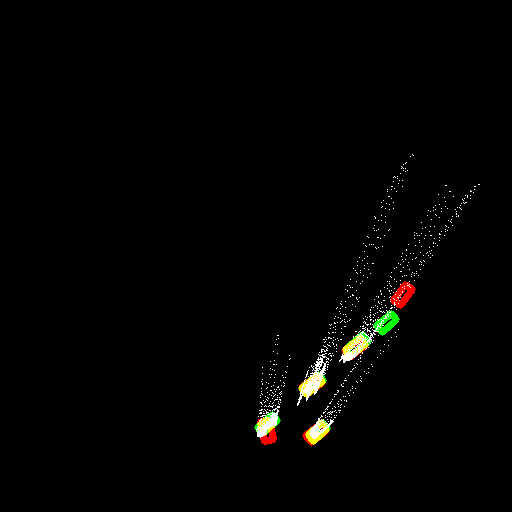}
    }
    \hfill
    \subfloat[3D detection using cuboids as object shapes.]{
        \includegraphics[width=0.32\linewidth,height=2cm]{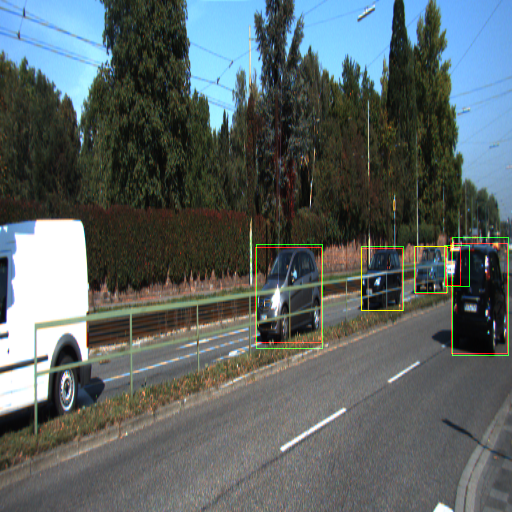}
        \includegraphics[width=0.32\linewidth,height=2cm]{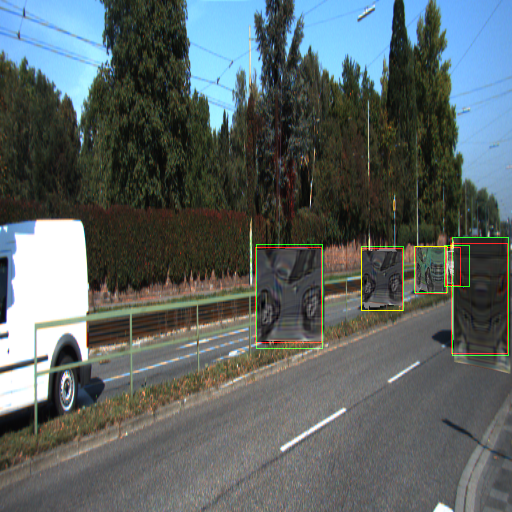}
        \includegraphics[width=0.32\linewidth,height=2cm]{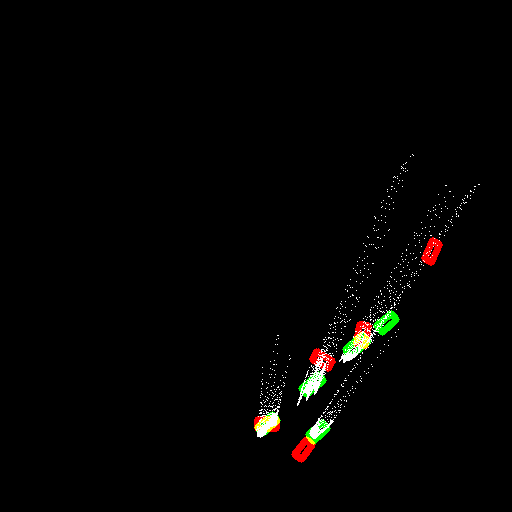}
    }
    \caption{The difference between 3D detection with and without shape optimization: Raw images with predicted 2D bounding boxes are shown in the left column. Reconstructed images (with the projection of an estimated 3D shape onto the image plane) are shown in the middle column. Ground-truth (green)/predicted (red) 3D bounding boxes in the bird's-eye view (BEV) are shown in the right column. The white points in BEV represent projections of object point clouds generated from predicted depth maps. The ground truth information is only used for visualization purposes.}
    \label{fig:shape_optimization}
\end{figure*}

Table~\ref{table:shape_optimization} shows a quantitative evaluation of these approaches. When the shape is not optimized, the detection accuracy drops by about 20\%-30\%. When cuboids are used as shapes, the detection accuracy decreases to nearly zero. Fig.~\ref{fig:shape_optimization} shows the difference between the two approaches. If the shapes are not optimized, the (projected) object silhouettes negatively impact the estimation of rotation, when using the render-and-compare approach. As we optimize for the silhouette loss along with fitting within the bounding box area, if the shapes are cuboids, optimization results in trying to fill the area inside the bounding boxes and silhouette loss becomes ineffective. These experimental results demonstrate that optimizing shapes is essential for our method to work effectively.

\begin{table}[!ht]
    \small
    \begin{center}
        \begin{tabular}{l|c|ccc|ccc}
            \toprule
            \multirow{2}{*}{Method} & \multirow{2}{*}{Supervised} & \multicolumn{3}{c|}{3D detection} & \multicolumn{3}{c}{Bird’s eye view} \\ 
             ~ & ~ &  Easy & Moderate & Hard  & Easy & Moderate & Hard \\ 
            \midrule   
            \midrule     
            Mono3D\cite{chen2016monocular} & \checkmark & 2.53 &  2.31 & 2.31 & 5.22 & 5.19 & 4.13  \\
            OFTNet\cite{roddick2018orthographic} & \checkmark & 4.07 & 3.27 & 3.29 & 11.06 & 8.79 & 8.91  \\
            FQNet\cite{liu2019deep} & \checkmark & 5.98 & 5.50 & 4.75 & 9.50 & 8.02 & 7.71 \\
            ROI-10D\cite{manhardt2019roi} & \checkmark & 9.61 & 6.63 &  6.29 & 14.50 & 9.91 & 8.73 \\
            Mono3D++\cite{he2019mono3d++} & \checkmark & 10.60 & 7.90 & 5.70 & 16.70 & 11.50 & 10.10 \\
            MonoGRNet\cite{qin2019monogrnet} & \checkmark & 13.88 & 10.19 & 7.62 & - & - & -   \\
            MonoDIS~\cite{simonelli2019disentangling} & \checkmark & 18.05 & 14.98 & 13.42 & 24.26 & 18.43 & 16.95  \\
            \midrule     
            MonoDR    & & 13.90  & 14.17  & 12.12  &  21.20 &  17.35 & 15.25 \\
            \bottomrule
        \end{tabular}
    \end{center}
    \caption{Evaluation of different monocular 3D detection methods: We report $AP_{|R_{11}}$ on the KITTI 3D {\it validation} set. The values are calculated assuming an intersection-over-union (IoU) between the predicted and ground truth bounding boxes of at least 0.7.}
    \label{table:val_ap11}
\end{table}
%%%%%%%% table %%%%%%%%%%%%%%%%%%%%%%%%

%%%%%%%% table %%%%%%%%%%%%%%%%%%%%%%%%
% Statistical on KITTI3D
\begin{table*}[h]
    \small
    \begin{center}
        \begin{tabular}{c|c|ccc|ccc}
            \toprule
            \multirow{2}{*}{Metric} & \multirow{2}{*}{Threshold} & \multicolumn{3}{c|}{3D detection} & \multicolumn{3}{c}{Bird’s eye view} \\
             ~ & ~ & Easy & Moderate & Hard  & Easy & Moderate & Hard \\ 
            \midrule
            $AP_{|R_{40}}$ & 0.7 & 12.50 &  7.34 &  4.98 & 19.49 & 11.51 & 8.72 \\
            $AP_{|R_{40}}$ & 0.5 & 43.37 & 29.50 & 22.72 & 48.53 & 33.90 & 25.85  \\
            $AP_{|R_{40}}$ & 0.3 & 65.58 & 52.02 & 42.38 & 68.62 & 54.94 & 45.29 \\
            $AP_{|R_{11}}$ & 0.7 & 18.86 & 14.04 & 12.05 & 24.79 & 17.10 & 15.01 \\
            $AP_{|R_{11}}$ & 0.5 & 45.76 & 32.31 & 26.19 & 51.13 & 37.29 & 30.20 \\
            $AP_{|R_{11}}$ & 0.3 & 66.44 & 52.09 & 44.30 & 68.94 & 57.00 & 45.66 \\
            \bottomrule
        \end{tabular}
    \end{center}
    \caption{Quantitative evaluation of different metrics and thresholds on the KITTI {\it validation} set.}
    \label{table:metrics_thresholds}
\end{table*}
%%%%%%%% table %%%%%%%%%%%%%%%%%%%%%%%%

%%%%%%%% table %%%%%%%%%%%%%%%%%%%%%%%%
% Is depth needed?
% TODO : Replace GeoNet with Monodepth2 
%\begin{table*}[t]
%    \small
%    \begin{center}
%        \begin{tabular}{l|ccc|ccc}
%            \toprule
%            \multirow{2}{*}{Depth Network} & \multicolumn{3}{c|}{3D detection} & \multicolumn{3}{c}{Bird’s eye view} \\ 
%            ~ & Easy & Moderate & Hard  & Easy & Moderate & Hard \\ 
%            \midrule   
%            GeoNet\cite{yin2018geonet}   & TODO & TODO & TODO & TODO & TODO & TODO \\
%            PackNet\cite{guizilini20193d}  & 12.50 &  7.34 & 4.98 & 19.49 & 11.51 & 8.72 \\
%            LIDAR point clouds  & TODO & TODO & TODO & TODO & TODO & TODO \\
%            \bottomrule
%        \end{tabular}
%    \end{center}
%\caption{Effect of the input depth on the accuracy: Average Precision ($AP_{loc}$) (in \%) of bird's eye view boxes on KITTI {\it validation} set.}
%    \label{table:depth_networks}
%\end{table*}
%%%%%%%% table %%%%%%%%%%%%%%%%%%%%%%%%

%%%%%%%% table %%%%%%%%%%%%%%%%%%%%%%%%
% "Is shape reconstruction important?"
\begin{table*}[!ht]
    \small
    \begin{center}
        \begin{tabular}{c|ccc|ccc}
            \toprule
            \multirow{2}{*}{Design choice} & \multicolumn{3}{c|}{3D detection} & \multicolumn{3}{c}{Bird’s eye view} \\
             ~ & Easy & Moderate & Hard  & Easy & Moderate & Hard \\ 
            \midrule
            Optimizing shapes & 12.50 &  7.34 & 4.98 & 19.49 & 11.51 & 8.72 \\
            Using random shapes & 9.10 & 5.51 & 4.01 & 16.28 & 10.26 & 7.48 \\
            Using cuboids & 0.30 & 0.17 & 0.16 & 0.79 & 0.48 & 0.44 \\
            \bottomrule
        \end{tabular}
    \end{center}
    \caption{Effectiveness of shape reconstruction on the KITTI {\it validation} set.}
    \label{table:shape_optimization}
\end{table*}
%%%%%%%% table %%%%%%%%%%%%%%%%%%%%%%%%

%% file: 08_acknowledgements.tex
\section{Acknowledgements}
This work was supported by Toyota Research Institute Advanced Development, Inc. The authors would like to thank Richard Calland, Karim Hamzaoui, Rares Ambrus, Vitor Guizilini for their helpful comments and suggestions.